\pdfoutput=1
\documentclass{article}

\usepackage[preprint]{neurips_2023}

\usepackage{natbib}
\setcitestyle{numbers,square}

\usepackage[utf8]{inputenc} 
\usepackage[T1]{fontenc}    
\usepackage[colorlinks=true, linkcolor=red, citecolor=green, urlcolor=blue]{hyperref}
\usepackage{url}            
\usepackage{booktabs}       
\usepackage{amsfonts}       
\usepackage{nicefrac}       
\usepackage{microtype}      
\usepackage{xcolor}         
\usepackage{amsmath}
\usepackage{amssymb}
\usepackage{wrapfig}
\usepackage{multirow}
\usepackage{graphicx}

\def\Datasetname{UniG3D}

\title{UniG3D: A Unified 3D Object Generation Dataset}

\newcommand*{\affaddr}[1]{#1}
\newcommand*{\affmark}[1][*]{\textsuperscript{#1}}

\newcommand*{\email}[1]{\texttt{#1}}

\author{
    Qinghong Sun\affmark[1]\thanks{Equal Contribution},
    Yangguang Li\affmark[1]\footnotemark[1], 
    ZeXiang Liu\affmark[1], 
    Xiaoshui Huang\affmark[2], 
    Fenggang Liu\affmark[1] \\
    \textbf{Xihui Liu}\affmark[3], 
    \textbf{Wanli Ouyang}\affmark[1], 
    \textbf{Jing Shao}\affmark[1]\thanks{Corresponding Author} \\
    \affaddr{\affmark[1]SenseTime Research}
    \affaddr{\affmark[2]Shanghai AI Lab} 
    \affaddr{\affmark[3]The University of Hong Kong}\\
    \small\email{\{sunqinghong1,liyangguang\}@sensetime.com}
}

\begin{document}
\maketitle

\begin{abstract}
The field of generative AI has a transformative impact on various areas, including virtual reality, autonomous driving, the metaverse, gaming, and robotics. Among these applications, 3D object generation techniques are of utmost importance. This technique has unlocked fresh avenues in the realm of creating, customizing, and exploring 3D objects.
However, the quality and diversity of existing 3D object generation methods are constrained by the inadequacies of existing 3D object datasets, including issues related to text quality, the incompleteness of multi-modal data representation encompassing 2D rendered images and 3D assets, as well as the size of the dataset.
In order to resolve these issues, we present \textbf{\Datasetname}, a unified 3D object generation dataset constructed by employing a universal data transformation pipeline on Objaverse and ShapeNet datasets. 
This pipeline converts each raw 3D model into comprehensive multi-modal data representation <text, image, point cloud, mesh> by employing rendering engines and multi-modal models. These modules ensure the richness of textual information and the comprehensiveness of data representation.
Remarkably, the universality of our pipeline refers to its ability to be applied to any 3D dataset, as it only requires raw 3D data. The selection of data sources for our dataset is based on their scale and quality. 
%
Subsequently, we assess the effectiveness of our dataset by employing Point-E and SDFusion, two widely recognized methods for object generation, tailored to the prevalent 3D representations of point clouds and signed distance functions.
%
%
\textbf{Our dataset is available at: \href{https://unig3d.github.io}{https://unig3d.github.io.}}

\end{abstract}

\section{Introduction}
\begin{figure}[t]
    \centering
    \includegraphics[width=1.0\textwidth]{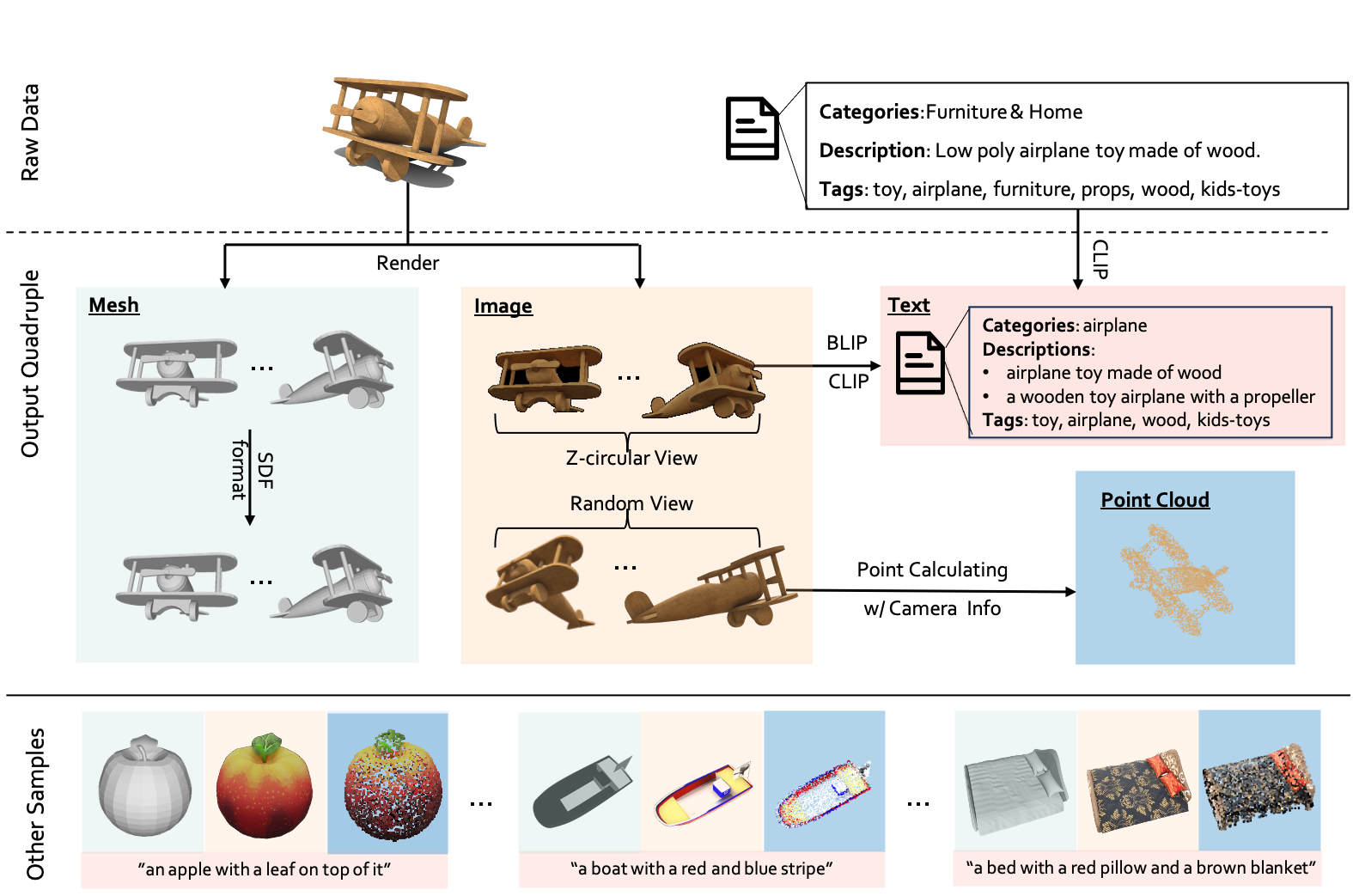}
    \vspace{-1em}
    \caption{Overview of the data  transformation pipeline of \Datasetname. First, we use Blender~\cite{kent20153d} to render each 3D model to multi-view meshes and images. Next, we obtain colored point clouds by calculating the points of multiple random-view rendered images. Finally, we employ the CLIP-VIT~\cite{radford2021learning} and multi-modal LLM~\cite{li2022blip}  to acquire rich text information. Additionally, we present some quadruples within the datasets.
    }
    \label{fig:fig2}
\end{figure}

Generative AI has revolutionized the way humans work and improved their efficiency, as this technology can understand human intentions and automatically generate the required content.
In particular, there have been numerous generative works in virtual reality~\cite{armeni20163d, liu2021group, misra2021end, vu2022softgroup}, autonomous driving~\cite{li2022deepfusion, yin2021center}, metaverse~\cite{lin2023blockchain, lin2023unified} and gaming~\cite{kim2020learning, torrado2018deep}, and robotics~\cite{wojek2011monocular, cadena2016multi}. In the aforementioned applications, a crucial and significant technique is 3D object generation~\cite{zeng2022lion, zhou20213d, jain2022zero, poole2022dreamfusion, lin2022magic3d, nichol2022point, cheng2022sdfusion, jun2023shapp}, which involves creating realistic or novel 3D representations aiming to simulate and replicate real-world or imagined 3D objects. 3D object generation technology opens up new possibilities for creating, customizing, and exploring 3D objects, making it a valuable tool in various fields where 3D models play a crucial role.

Several recent works~\cite{jain2022zero, poole2022dreamfusion, lin2022magic3d} have tackled the problem of 3D generation by optimizing 3D representations against a text-to-image model and do not leverage 3D data. Although these methods have demonstrated promising results, state-of-the-art approaches typically require about six GPU hours to produce a single sample. It is challenging to scale up the generation of 3D data by utilizing these methods. 
There are alternative methods for 3D object generation that make use of 3D data. Some methods incorporate text as a condition during the 3D generation process.
Despite promising early results, many of these works are limited to simple prompts or a narrow set of object categories due to the scarcity of 3D training data~\cite{liu2022towards, sanghi2022clip, zeng2022lion, mittal2022autosdf, fu2022shapecrafter, cheng2022sdfusion}. 
Alternatively, some methods use a pre-trained text-to-image model to condition their 3D generation procedure~\cite{nichol2022point, jun2023shapp}. However, since most datasets lack image data, researchers are left with the task of rendering each dataset individually, which is a time-consuming and resource-intensive process. In addition to text-conditioning, existing methods also employ image-conditioning as an alternative approach. However, they face similar challenges as mentioned above.
As a result, the quality of text information, the availability of 2D-rendered data, and the scalability of the dataset are crucial. 

To resolve these issues, we construct a unified 3D object generation dataset, \textbf{\Datasetname}, by utilizing the ShapeNet~\cite{shapenet} and Objaverse~\cite{deitke2022objaverse} as raw data sources. We develop a \emph{unified} pipeline that can convert raw 3D models into comprehensive multi-modal data, which allows researchers focusing on different target 3D representations or different input conditions to use it conveniently. Specifically, as shown in Fig.~\ref{fig:fig2}, we convert each raw 3D model into a <text, image, point cloud, mesh> quadruple by employing a 3D model rendering engine~\cite{kent20153d}, BLIP~\cite{li2022blip}, and CLIP~\cite{radford2021learning} model. The former is used for rendering 2D and 3D representations, while the latter two are used for generating high-quality textual information.
Our proposed unified multi-modal data  transformation pipeline requires only 3D data, which eliminates the need for any manual annotation effort and demonstrates its scalability. To illustrate its effectiveness, we conduct our dataset by using ShapeNet~\cite{shapenet} and Objaverse~\cite{deitke2022objaverse} as the data sources, given their scale and quality. Then, we validate the efficacy of our dataset by utilizing Point-E~\cite{nichol2022point} and SDFusion~\cite{cheng2022sdfusion}, two typical object generation methods that are widely used in prevalent 3D representations such as point cloud and signed distance function.
\Datasetname~offers three contributions:
\begin{enumerate} 
\item We construct a large-scale unified 3D object generation dataset with rich textural information and comprehensive multi-modal data.
\item We propose a universal data transformation pipeline that can convert any 3D data into representations suitable for most 3D object generation methods.
\item To validate the efficacy of our dataset, we conduct experiments under various input conditions and target 3D representations. Based on our empirical investigations, we present several valuable insights into the impact of various conditions, the efficacy of data expansion, and the significance of text quality.

\end{enumerate}

\section{Related Work}

\subsection{3D Generative Methods}
Several recent works have explored the challenge of generating 3D models with conditioned inputs by optimizing the 3D representations based on a text-image matching objective. \cite{jain2022zero} introduce DreamFields, a method that leverages CLIP to optimize the parameters of a NeRF model without the need for 3D training data. More recently, \cite{poole2022dreamfusion} extends DreamFields by incorporating a pre-trained text-to-image diffusion model in place of CLIP, producing more coherent and complex objects. \cite{lin2022magic3d} builds upon this technique by converting the NeRF representation into a mesh and further refining the mesh representation through a secondary optimization stage. Although these approaches are capable of generating diverse and intricate objects or scenes, the optimization procedures often demand significant GPU computational time to converge, posing challenges for practical applications.

While the above primarily rely on optimizing against a 2D text-image model and do not utilize 3D data, alternative methods for conditional 3D object generation incorporate 3D data, sometimes in conjunction with text labels. \cite{liu2022towards} leverages paired text-3D data to generate models in a joint representation space. \cite{sanghi2022clip} employs a flow-based model to generate 3D latent representations, and find some text-to-3D capabilities when conditioning their model on CLIP embeddings. 
\cite{mittal2022autosdf} and \cite{fu2022shapecrafter} employ a VQ-VAE with an autoregressive prior to sample 3D shapes conditioned on text labels. SDFusion~\cite{cheng2022sdfusion} utilizes an encoder-decoder structure to compress 3D shapes into a compact latent representation, which is then used to train a diffusion model for text-to-3D generation. While many of these works demonstrate promising early results, they tend to be limited to simple prompts or a narrow set of object categories due to the limited availability of 3D training data. Point-E~\cite{nichol2022point} solves this problem by building a large-scale text 3D dataset. However, the datasets are not open source. 

Alternatively, some methods ~\cite{jun2023shapp, nichol2022point}use a pre-trained text-to-image model to condition their 3D object generation procedure with images. However, since most datasets lack image data, researchers are left with the task of rendering each dataset individually, which is a time-consuming and resource-intensive process. In addition to text-conditioning, existing methods also employ image-conditioning as an alternative approach. However, they face similar challenges as mentioned above.

\subsection{3D Object Datasets}
Many widely-used 3D datasets prefer to collect synthetic CAD models from online repositories~\cite{shapenet, modelnet40, 3d_future, Abo, deitke2022objaverse}. Shapenet\cite{shapenet} stands out as the prevalent dataset. It covers 55 common object categories with about 51,300 unique 3D models. Every object in this dataset is precisely rendered as a 3D mesh, thereby imparting meticulous geometric information. Moreover, a "name" field supplements each 3D model, carrying rich metadata associated with it. Objaverse~\cite{deitke2022objaverse} is an extensive dataset comprising over 800K 3D models accompanied by descriptive captions, tags, and animations. It surpasses existing 3D repositories in terms of its scale, the number of categories, and the visual variety of instances within each category. However, the majority of objects lack appropriate text information. Each object in the dataset is represented in GLB format, posing a challenge for users in directly leveraging the 3D data.
Another line of works~\cite{DTU, BlendedMVS, ScannObjectNN, ScannObjectNN, GSO, AKB-48, co3d, OmniObject3D, MVImgNet, YCB, Bigbird} advocate real 3D objects in limited scale. MVImgNet~\cite{MVImgNet} is a recent medium-sized dataset of multi-view images, which is highly convenient to gain by shooting videos of real-world objects in human daily life. It contains 6.5 million frames from 219,188 videos crossing objects from 238 classes, with rich annotations of object masks, camera parameters, and point clouds. 

As previously discussed, the current state of 3D object generation techniques is hindered by various factors that limit their quality. These factors include deficiencies in text quality, limited availability of 2D-rendered data, and dataset scale. Apart from the dataset scale, these limitations manifest in two ways. Firstly, available text information is often imprecise and lacking in detail, providing only broad categories or incomplete descriptions that have limited correlation with actual models. 
Additionally, existing datasets typically lack the essential data formats required for 3D object generation tasks, such as multi-view 2D rendered images, 3D point clouds, and multi-view meshes. For example, Shapenet~\cite{shapenet} only provides raw meshes, while Objaverse~\cite{deitke2022objaverse} exclusively offers raw 3D models in GLB format. 

\section{The \Datasetname~Dataset}

\begin{wraptable}{r}{0.6\textwidth}
    \centering
    \vspace{-1em}
    \caption{The statistical information of the four representations in two \Datasetname~datasets.}
    \vspace{1em}
    \label{table:tab1}
    \resizebox{1\linewidth}{!}{
    \begin{tabular}{lcccc}
    \toprule
    \textbf{Dataset} & \textbf{Mesh} & \textbf{PCL} & \textbf{Image} & \textbf{Text} \\
    \midrule
    \Datasetname-Shapenet & 500K & 50K & 1 million & 50K  \\
    \Datasetname-Objaverse & 5 million & 500K & 10 million & 500K \\
    \bottomrule
    \end{tabular}
    }
\end{wraptable}

In this section, we describe the data transformation process of \Datasetname. The raw 3D data and related text information are gathered from two 3D datasets, specifically chosen among various datasets based on their scale and significance. The first dataset is Shapenet~\cite{shapenet}, a classic dataset that has around 50K+ 3D objects with 55 annotated categories. The second one is Objaverse~\cite{deitke2022objaverse}, the large-scale 3D dataset. It has approximately 800K 3D objects, but the category of most objects is unknown. We describe the transformation pipeline of different representations in more detail in the following. As shown in Fig.\ref{fig:fig2}, we convert each 3D model to four representations, which are mesh, image, point cloud, and rich text information. 

\subsection{The Construction of Quadruples}
\label{The Construction of Quadruples}
As illustrated in Fig.~\ref{fig:fig2}, we present the unified data transformation pipeline employed in our dataset. This pipeline encompasses the transformation of raw 3D data and raw text information into a unified representation, quadruple. The pipeline begins by leveraging a powerful 3D model rendering engine, which enables the generation of multiple 2D and 3D representations from the raw data. 
Furthermore, our pipeline incorporates the utilization of state-of-the-art multi-modal models to generate high-quality textual information. By employing these models, we extract meaningful and descriptive text features from the 3D representations. This process ensures that our dataset includes comprehensive and accurate textual information that complements the visual aspects of the objects.

\textbf{Mesh. }
For each 3D model, we employ Blender~\cite{kent20153d}, a versatile software tool that supports various 3D formats and incorporates an optimized rendering engine, to generate ten multi-view meshes in the OBJ format. These meshes are rendered using a z-circular camera pose, ensuring comprehensive coverage of the object from different viewpoints. Specifically, the views are evenly spaced at intervals of 36 degrees to capture a wide range of perspectives. Additionally, to facilitate model training~\cite{cheng2022sdfusion}, we also provide the models in signed distance function~(SDF) format, which offers a convenient representation for 3D object generation tasks.

\textbf{Image.}
To obtain multi-view 2D images for each 3D model, we implement a customized rendering process. Leveraging the capabilities of Blender~\cite{kent20153d}, we develop a script that first normalizes the 3D models to fit within a bounding cube, ensuring consistent scale across all objects. Additionally, we set up a standardized lighting arrangement to ensure uniform illumination across the rendered images.
Subsequently, we employ Blender's built-in real-time rendering engine to export the 2D images. The rendering process involved capturing ten images using the z-circular camera pose, which provides a well-distributed set of views around the object. These views are captured at equal intervals of 36 degrees, enabling comprehensive coverage of the object from different angles. The deflection angle used for these random poses is the same as that of the above-mentioned multi-view meshes.
In addition to the z-circular camera pose, we also capture another set of ten images using random camera poses. These random poses introduce variations in the viewing angles and orientations of the object, allowing for the generation of dense point clouds.
By combining the multi-view images captured using the z-circular camera pose and the random poses, we obtain a total of 20 multi-view 2D images for each 3D model. This diverse set of images provides comprehensive visual information for subsequent experiments and processing tasks.

\textbf{Point Cloud.}
To convert the 3D models into colored point clouds, we utilize the RGBAD images rendered from them. Initially, we generate dense point clouds by associating points with each pixel in the rendered RGBAD images. However, these point clouds often exhibit uneven distribution and contain a large number of points. To address this issue, we employ voxel point sampling techniques to create uniform point clouds consisting of 4K points.
By directly constructing point clouds from the rendered images, we circumvent potential challenges that may arise when sampling points from 3D meshes. These challenges include dealing with points located inside the model or handling 3D models stored in non-standard file formats~\cite{nichol2022point}. Our approach ensures a consistent and reliable representation of the objects in the form of point clouds.
To further enhance the quality of our dataset, we implement heuristics to exclude low-quality models. Specifically, we employ a criterion based on the singular value decomposition (SVD) of each point cloud~\cite{nichol2022point}. We compute the SVD for each point cloud and retain only those models where the smallest singular value exceeds a certain threshold. This process effectively filters out flat objects or models with poor geometric structure, ensuring that our dataset comprises high-quality and meaningful 3D representations.
By employing these techniques, we create a comprehensive dataset of colored point clouds that accurately represent the underlying 3D models, while also ensuring the inclusion of high-quality and diverse objects for further analysis and research.

\textbf{Text.}
The text information of the 3D model mainly has two sources. One is the raw text information associated with each 3D model. However, we observe that a significant portion of this text information does not accurately correspond to the 3D models themselves, including terms like "model", "blender", and "Low poly". Therefore, we employ CLIP-VIT model~\cite{radford2021learning} to clean them by calculating image-text similarity. We only retain those where the similarity value was above a certain threshold. By employing this method, over 80\% of low-quality texts can be filtered out, while the false recognition rate does not exceed 30\%. The other source is employing multi-modal LLM to generate descriptions with its 2D image as input. For each 3D model, we employ BLIP~\cite{li2022blip} to generate rich and detailed descriptions based on its thumbnail or 2D rendered image. Then, we also evaluate its accuracy by the similarity score using CLIP-VIT model~\cite{radford2021learning}. 
\begin{wrapfigure}{r}{0.5\textwidth}
    \centering
    \includegraphics[width=0.5\textwidth]{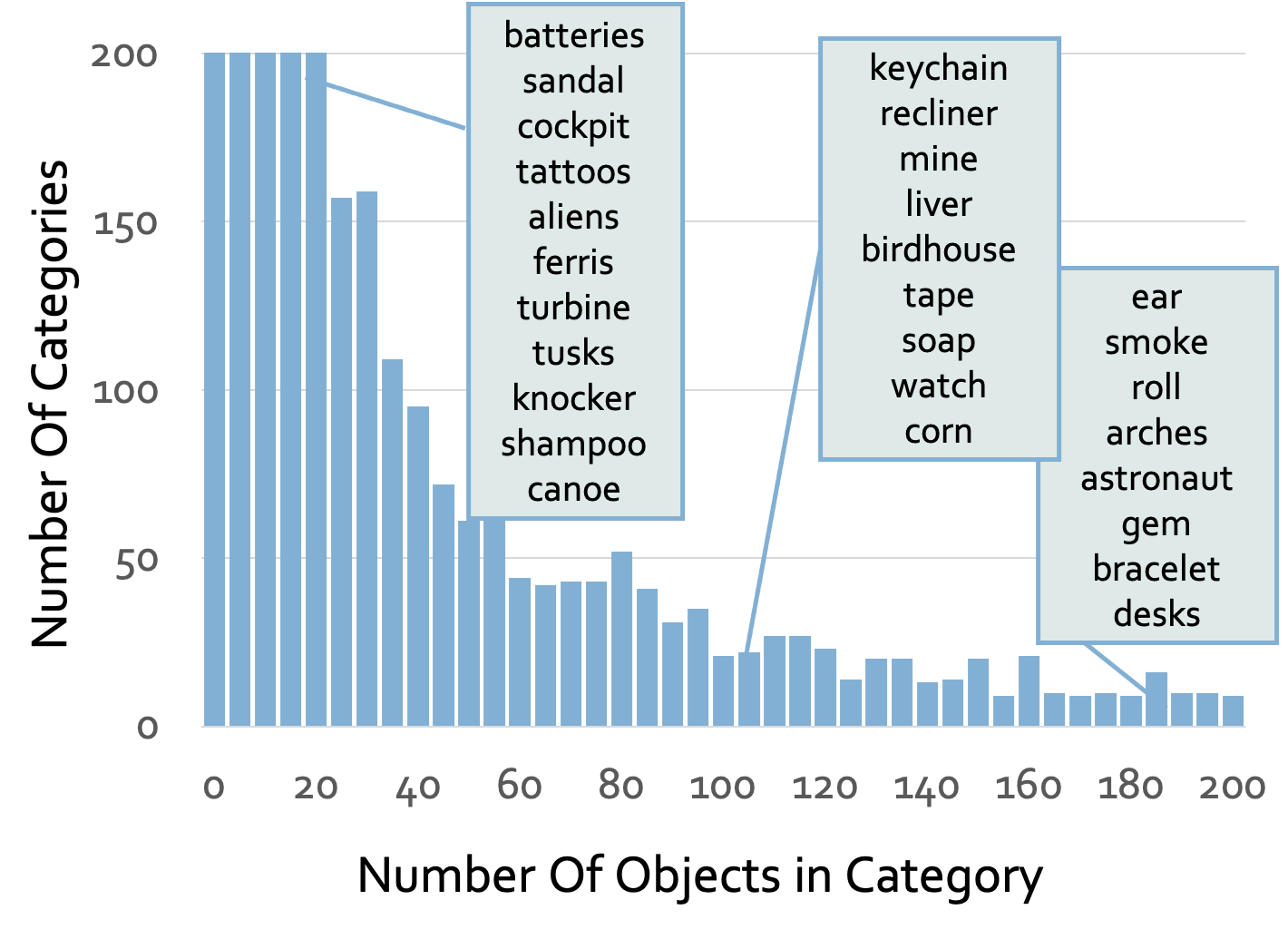}
    \caption{
    A histogram of fine-grained \Datasetname~categories with representative members from several bins highlighted.}
\label{fig:dataset}
\end{wrapfigure}
Based on our data observation, we find that the data with low similarity is primarily attributed to the abstraction of 3D models, which have weak visual features. Hence, we implement a filtering process where we retain only those 3D models whose similarity score surpasses a predetermined threshold. By employing this method, the models we filter out account for approximately 20\% of the total. Furthermore, for the 3D models with known categories, we enhance their description information by aligning it with the corresponding category. To be specific, if the category sentence is absent from the original description, we extract descriptive phrases from the existing description and merge them with the known category sentence to create an improved description. Conversely, if the generated description already exhibits high quality and coherence, we retain it as is.

\subsection{Statistics of our dataset}

The statistics for each representation of \Datasetname~are presented in Table \ref{table:tab1}, providing a significant overview. Our dataset pipeline generates multiple representations for each raw 3D model, including ten meshes, one colored point cloud, and 20 images. These visual representations are accompanied by corresponding descriptive text information, enhancing the richness and comprehensiveness of the dataset.
To provide a visual overview of the dataset, Fig.~\ref{fig:dataset} showcases the varying degrees of object counts across different categories within \Datasetname, highlighting both the long-tail and head categories and their respective object counts.
The distribution of object counts in each category ranges from 0 to 200 in our dataset. More than half of the categories have fewer than 20 objects, representing the long-tail categories of the dataset. In contrast, there are approximately 50 head categories that contain around 200 objects each, representing the head categories.

\section{Experiments}

\subsection{3D Object Generation Method}
\begin{wrapfigure}{r}{0.5\textwidth}
    \vspace{-2em}
    \centering
    \includegraphics[width=0.5\textwidth]{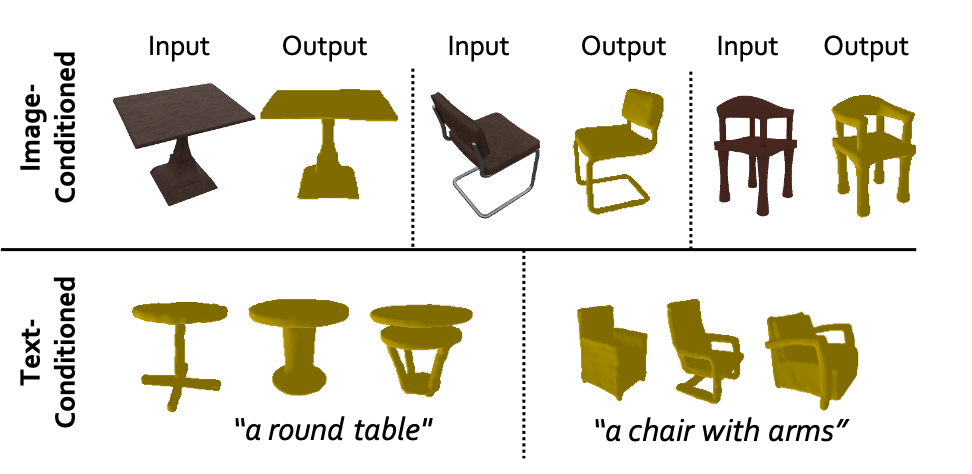}
    \caption{
    Mesh generated by SDFusion conditioned on different input modalities of Shapenet.}
\label{fig:conds_text-img}
\end{wrapfigure}
\begin{figure}[t]
    \centering
    \includegraphics[width=1.0\textwidth]{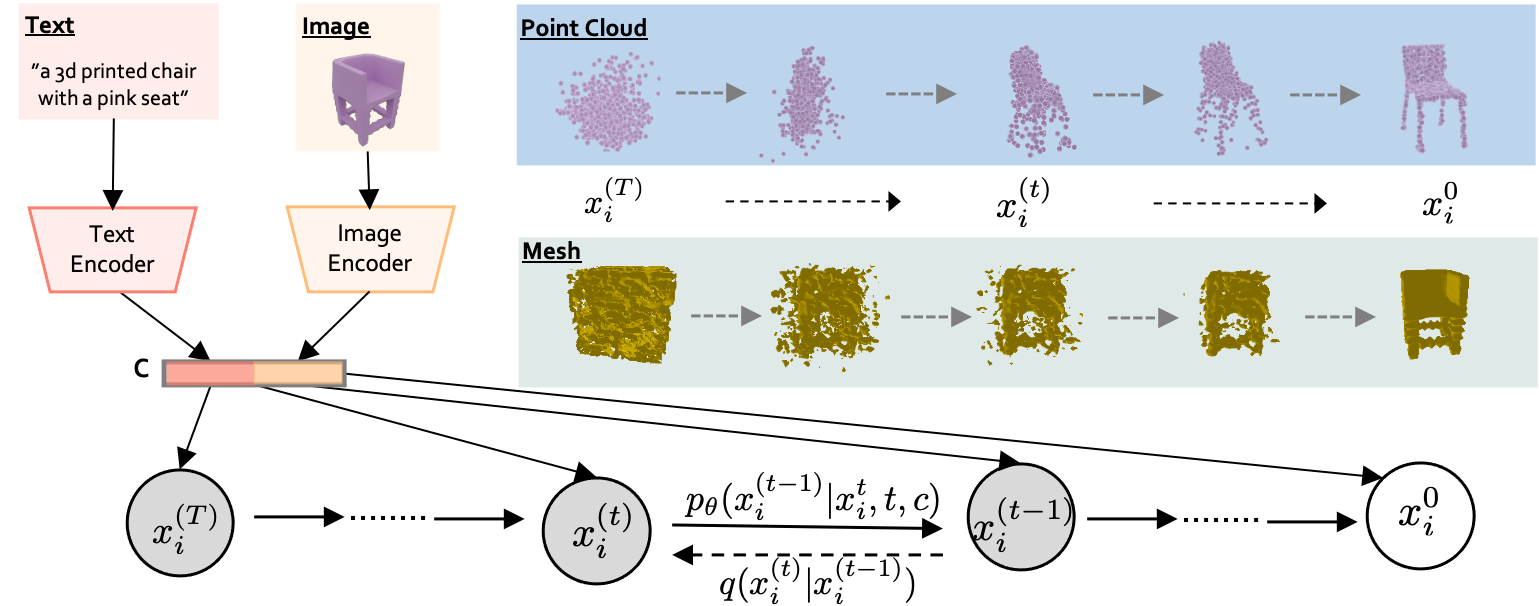}
    \vspace{-1em}
    \caption{The directed graphical model depicts the diffusion process for 3D representations. The variable $x_i^{(t)}$ denotes the 3D representation at timestep $t$. The processes of noise addition and denoising are represented by $q$ and $p$ respectively. The denoising process is conditioned on variable $c$, which can be either embedded text, embedded image, or a combination of both.
    }
\label{fig:difussion_process_compressed}
\end{figure}

In our experimental study, we leverage two commonly used 3D object generation methods, namely Point-E and SDFusion, which are specifically tailored for two prevalent 3D representations, which are point cloud and signed distance function. These methods are grounded in the diffusion process, as proposed by Sohl-Dickstein et al.~\cite{sohl2015deep}. To enhance the clarity of the training and generation processes, we depict the forward and backward processes in Fig. \ref{fig:difussion_process_compressed}, which provides a comprehensive overview of the steps involved in both the forward and backward pass of the diffusion model. Please refer to the supplementary material for detailed hyperparameters of the training process.
In particular, we utilize two model structures in SDFusion: VQ-VAE and the 3D latent diffusion model, whose parameter count exceeds 400 million. To provide the ability for interaction, learning conditional distribution is important. We incorporate multiple conditional input modalities with task-specific encoders and cross-attention modules in the latent diffusion model, such as text input, image input, and text-image multi-modality input. For text input, we embed the text caption using BERT~\cite{devlin2018bert}, while for image input, we embed the image using CLIP~\cite{radford2021learning}.
As for the generation of the point cloud, we employ two small model structures in Point-E, which are 40M-text and 40M-image. Specifically, 40M-text is a small model which only conditions text captions, not rendered images. The text caption is embedded with CLIP, and the CLIP embedding is appended as a single extra token of context. This model depends on the text captions present in our 3D dataset and does not leverage the fine-tuned GLIDE model. 40M-image is a small model with full image conditioning through a grid of CLIP latent. In the future, we will expand the scale of the training dataset and model structure.

\subsection{Experimental Results}

\subsubsection{The effects of different conditions}

\begin{wrapfigure}{r}{0.5\textwidth}
    \vspace{-1em}
    \centering
    \includegraphics[width=0.5\textwidth]{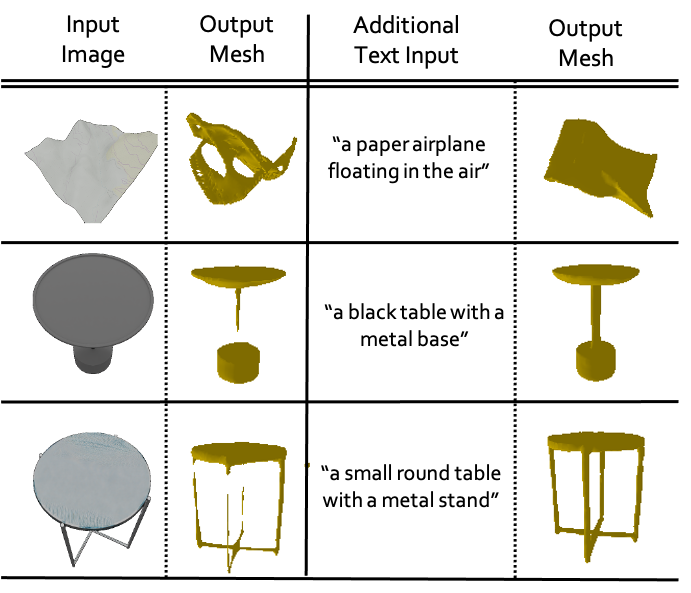}
    \caption{
    Comparison of the generative results by SDFusion when using images as the sole condition versus when incorporating additional text information.}
\label{fig:conds-supply}
\end{wrapfigure}

To explore the distinct roles of different input modalities in the 3D object generation process, we initially compare the effects of images and text. The experiment is conducted on Shapenet. As shown in Fig.\ref{fig:conds_text-img}, images can provide a direct visual reference, allowing for the generation of 3D objects that closely resemble the appearance of the referenced images. However, images may lack contextual information or high-level semantics that can be conveyed through textual descriptions, as shown in the second column of Fig.~\ref{fig:conds-supply}. 
Textual descriptions allow for precise and specific control over the generated 3D objects by providing detailed instructions or constraints. But, textual descriptions may sometimes be ambiguous or subjective, leading to different interpretations and potential variations in the generated 3D objects. Additionally, text lacks the ability to convey rich visual details, such as color gradients, textures, or fine-grained shape features.

Consequently, using text and images as conditioning inputs have its own advantages and disadvantages. Moreover, we aim to explore the effect when utilizing both of the aforementioned modalities as conditions. As shown in Fig.\ref{fig:conds-supply}, when using images as the sole condition, the model may have limited semantic understanding due to issues such as the angle of the images. For example, in the first row, the paper airplane, as well as the chairs in the second and third rows, can be challenging for the model to accurately determine the specific category or infer the occluded parts solely based on the image information.
By incorporating text as an additional modality, the model gains a better grasp of the desired object characteristics, resulting in improved generation quality and a more comprehensive understanding of the semantic information associated with the object.
Overall, using both text and images as conditioning inputs in 3D generation methods offers complementary benefits, with text providing semantic control and language understanding, while images contribute visual realism and rich visual details. The choice between these modalities depends on the specific requirements and objectives of the 3D generation task.

\begin{wrapfigure}{r}{0.5\textwidth}
    \vspace{-2em}
    \centering
    \includegraphics[width=0.5\textwidth]{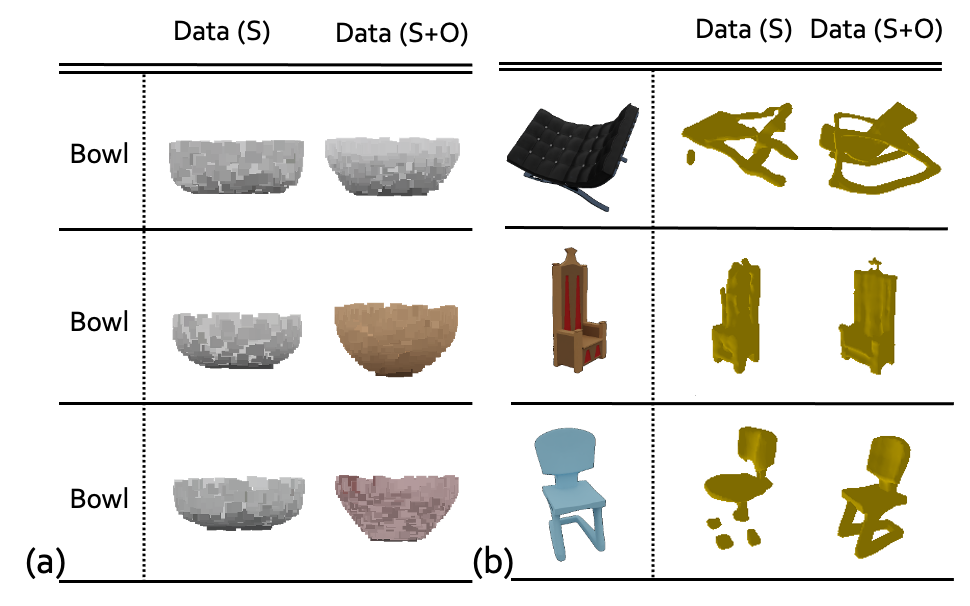}
    \caption{Effects of increased data sources. Data(s) represents the use of only the ShapeNet, while Data(s+o) represents the additional inclusion of the \Datasetname-Objaverse$_{\text{coco}}$ dataset. (a) and (b) represent the experiments conducted on different 3D representations.}
\label{fig:data-supply}
\end{wrapfigure}

\subsubsection{The effectiveness of data expansion}

Due to limited computing resources, we are currently unable to conduct experiments with the entire set of \Datasetname-Objaverse datasets. Therefore, we create a subset of the dataset by selecting data from the coco category, which we refer to as \Datasetname-Objaverse$_{\text{coco}}$. This subset contains around 50K 3D objects with 66 categories. \Datasetname-Shapenet has 50K+ 3D objects with 55 categories.

In order to explore the changes in data generation quality, such as point cloud integrity and diversity, after incrementally adding data from different categories, we conduct experiments on two 3D representations. 
Firstly, due to the high training cost associated with incorporating images as conditions in the Point-E method, we conduct experiments initially using text as the conditioning modality. As shown in Fig.~\ref{fig:data-supply} (a), increasing the sources of data enhances the diversity of generated 3D models. This implies the necessity for large-scale and scalable datasets. Due to the model's sensitivity to language ambiguity and variation, it is important to note that when integrating different data sources, language ambiguities need to be addressed. For example, in different datasets, the term "mouse" could refer to either a small rodent or a computer input device. 
Then, we explore the impact of data expansion under different representations and conditions based on SDFusion method. The first column in Fig.~\ref{fig:data-supply} (b) represents the test data from \Datasetname-Objaverse$_{\text{coco}}$. We observe that the model trained solely on Shapenet exhibits generalization ability. However, the addition of supplementary data allows the model to perform better on the new domain. 

\subsubsection{The impact of multi-view data}

\begin{wrapfigure}{r}{0.5\textwidth}
    \vspace{-1em}
    \centering
    \includegraphics[width=0.5\textwidth]{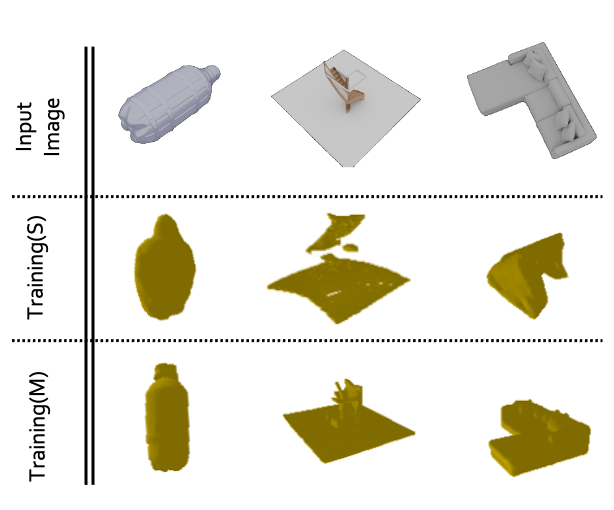}
    \vspace{-2em}
    \caption{Effects of increased multi-view data. Training(S) represents the model trained with single-view data, while Training(M) represents the model utilizing multi-view data.}
\label{fig:multi-view}
\end{wrapfigure}

In order to address the potential limitations of images captured from a single viewpoint, our dataset offers users multi-view data to facilitate data augmentation. As we describe in Section~\ref{The Construction of Quadruples}, our dataset includes ten multi-view rendered images along with their corresponding meshes. Specifically, the views are evenly spaced at intervals of 36 degrees to capture a wide range of perspectives.
To investigate the influence of multi-view data, we compare the performance of generative models trained on single-view and multi-view data using the SDFusion method. Training data is the combination of ShapeNet and \Datasetname-Objaverse$_{\text{coco}}$ dataset.
The benefits of utilizing multi-view training data are demonstrated in Fig.~\ref{fig:multi-view}, where it is evident that the model trained with such data outperforms in cases where the input image exhibits relatively uncommon viewpoints. Consequently, augmenting the dataset with multi-view perspectives proves to be an effective strategy for enhancing the model's robustness when dealing with less frequent angles.

\begin{wrapfigure}{r}{0.5\textwidth}
    \vspace{-1em}
    \centering
    \includegraphics[width=0.5\textwidth]{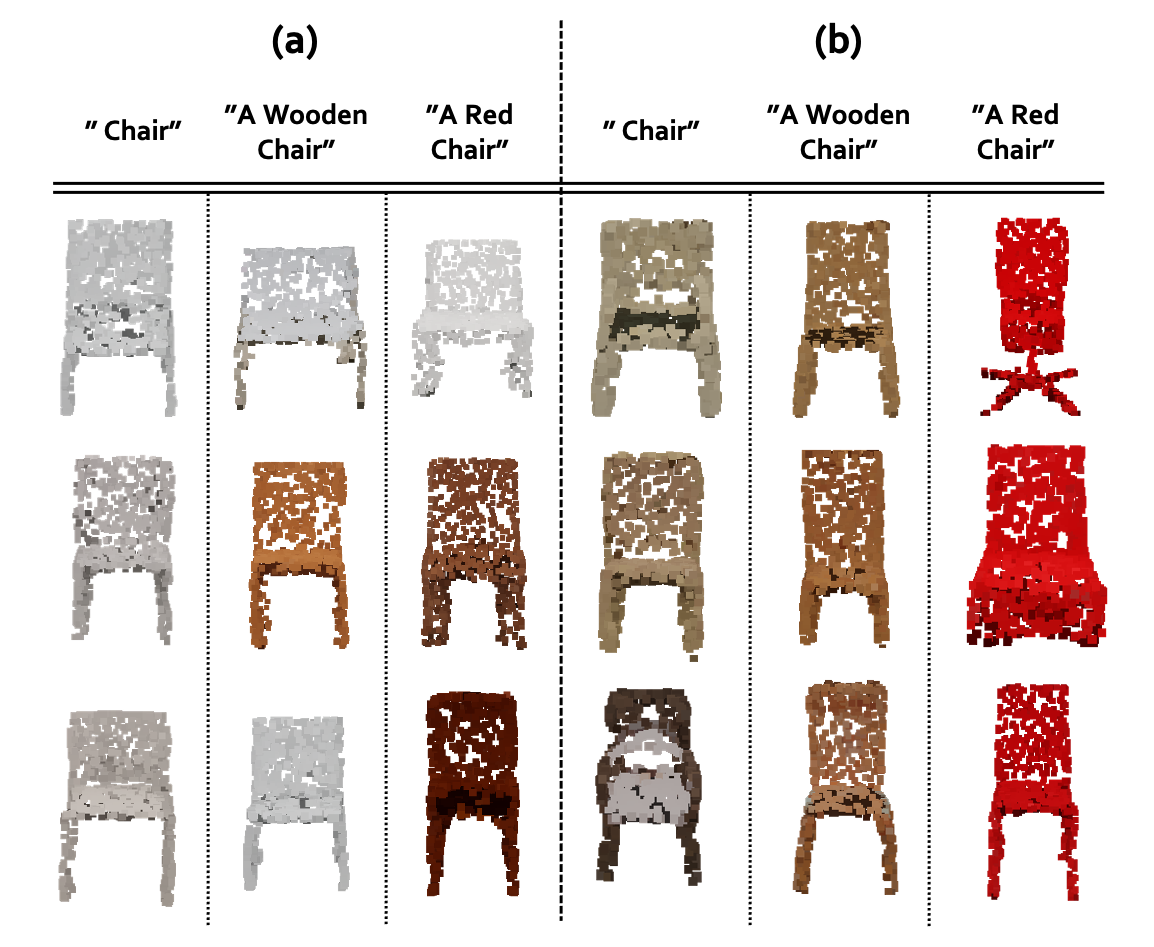}
    \caption{
    (a) corresponds to using only the category as text information during training, while (b) corresponds to utilizing the descriptive text generated by our pipeline.}
    \vspace{-1em}
\label{fig:fig4}
\end{wrapfigure}

\subsubsection{The importance of text quality}

Due to the relatively lower training cost of the generation experiments based on signed distance function representations, we employ Point-E to investigate the impact of text quality on the task of 3D object generation. To ensure data diversity, we utilize both the Shapenet and \Datasetname-Objaverse$_{\text{coco}}$ datasets as our training data. To provide a clear illustration of the quality and diversity of the generated point clouds, we present three different results for each text condition in Fig.\ref{fig:fig4}, using distinct seeds. 
It shows that when using category as the text condition in the inference, there is not much difference in the completeness and diversity of the generated point clouds. The model using category as its text information can only generate a chair model by inputting the word "chair". When the model is given descriptive text as input, it fails to generate results that possess matching visual features.
However, the model which has better text quality is capable of accepting more detailed descriptive text. 
It can generate more controllable models by specifying detailed textual information, such as specifying the airplane model's type, color, or material. 
Overall, our \Datasetname~data transformation pipeline enables the text-conditioned models to receive more detailed textual information and provide more control.

\section{Limitation and Social Impact}
Due to current limitations in computing resources, we are unable to conduct experiments using the complete \Datasetname-Objaverse datasets. However, in future work, we plan to present experimental results based on the entire dataset to validate the consistency of our conclusions on a larger scale. Moreover, considering the advancements in recent methods that demonstrate improved speed and quality, we intend to explore a broader range of 3D generation methods in our future experiments. Additionally, we aim to incorporate additional tasks related to 3D understanding, such as novel view synthesis, neural surface reconstruction, and 3D point cloud classification, to further expand the scope and applicability of our dataset.

This work aims to provide a unified dataset for the 3D generation task, eliminating the need for extensive human annotation efforts. While this approach offers positive impacts such as reducing human labor, it is crucial to acknowledge that the reduction of human labor may have negative consequences, including job loss or displacement, especially for individuals with lower skill levels who may rely on employment opportunities. 
\section{Conclusion}

In this study, we provide a unified 3D object generation dataset called \Datasetname. Our dataset is constructed by utilizing a universal data transformation pipeline applied to the Objaverse and ShapeNet datasets. This pipeline effectively converts each raw 3D model into a comprehensive multi-modal data representation, encompassing text, images, point clouds, and meshes. To achieve this, it utilizes rendering engines and multi-modal models that are capable of capturing textual information and ensuring a comprehensive representation of the data. As a result, our dataset guarantees the richness of textual information and the comprehensiveness of data representation.
Our pipeline's universality stems from its ability to be applied to any 3D dataset, solely relying on raw 3D data, thereby enhancing its practicality and flexibility. During the construction of our dataset, we meticulously select data sources considering their scale and quality, guaranteeing the incorporation of diverse and reliable information.
Furthermore, through empirical investigations and analysis of various factors, we present several key insights.

{\small
\bibliographystyle{unsrt}
\bibliography{Sections/egbib}
}

\clearpage

\end{document}